\documentclass[10pt,twocolumn,letterpaper]{article}

\usepackage[pagenumbers]{wacv} 
\usepackage[accsupp]{axessibility}
\usepackage{graphicx}
\usepackage{amsmath}
\usepackage{amssymb}
\usepackage{booktabs}
\usepackage{multirow}
\usepackage{url}
\usepackage{cite}
\usepackage{microtype}
\usepackage[normalem]{ulem}
\useunder{\uline}{\ul}{}
\usepackage{algorithm}
\usepackage{algpseudocode}
\usepackage{enumitem}
\usepackage{flushend}

\usepackage[pagebackref,breaklinks,colorlinks]{hyperref}

\usepackage[capitalize]{cleveref}
\crefname{section}{Sec.}{Secs.}
\Crefname{section}{Section}{Sections}
\Crefname{table}{Table}{Tables}
\crefname{table}{Tab.}{Tabs.}

\def\wacvPaperID{286} 
\def\confName{WACV}
\def\confYear{2026}

\newcommand{\pos}{\mathbf{y}}

\newcommand{\pixel}{\mathbf{y}}

\newcommand{\gdesc}{\mathbf{g}}
\newcommand{\ldesc}{\mathbf{l}}
\newcommand{\agg}{f_\text{agg}}

\newcommand{\diml}{\text{m}}
\newcommand{\dimg}{\text{n}}

\newcommand{\crd}{\mathbf{x}}

\newcommand{\image}{\mathit{I}}

\newcommand{\extrinsic}{H}

\begin{document}

\title{Robust Scene Coordinate Regression via\\ Geometrically-Consistent Global Descriptors}


\author{Son Tung Nguyen \qquad Alejandro Fontan \qquad Michael Milford \qquad Tobias Fischer\\[0.25cm]
Queensland University of Technology\\
Brisbane, Australia\\
{\tt\small sontung.nguyen@hdr.qut.edu.au}
}

\maketitle

\begin{abstract}

Recent learning-based visual localization methods use global descriptors to disambiguate visually similar places, but existing approaches often derive these descriptors from geometric cues alone (e.g., covisibility graphs), limiting their discriminative power and reducing robustness in the presence of noisy geometric constraints. We propose an aggregator module that learns global descriptors consistent with both geometrical structure and visual similarity, ensuring that images are close in descriptor space only when they are visually similar and spatially connected. This corrects erroneous associations caused by unreliable overlap scores. Using a batch-mining strategy based solely on the overlap scores and a modified contrastive loss, our method trains without manual place labels and generalizes across diverse environments. Experiments on challenging benchmarks show substantial localization gains in large-scale environments while preserving computational and memory efficiency. Code is available at \href{https://github.com/sontung/robust\_scr}{github.com/sontung/robust\_scr}.
 
 \end{abstract}
\begin{figure}[t]
    \centering
    \includegraphics[width=\linewidth]{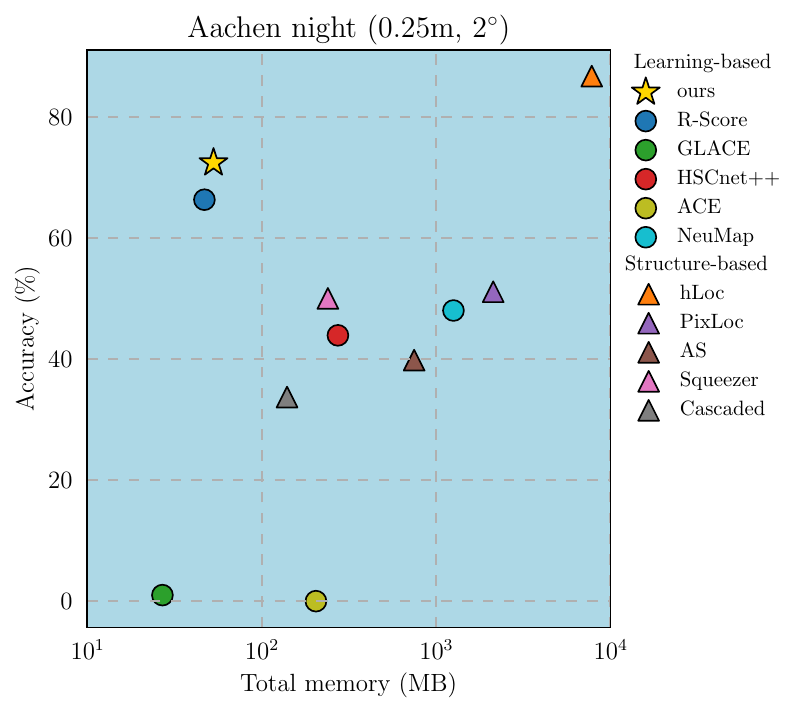}
        \vspace*{-4mm}
    \caption{\textbf{Performance overview on Aachen Day/Night dataset.} Our method achieves significant improvements over existing learning-based approaches while maintaining comparable memory efficiency. Compared to R-Score~\cite{rscore}, we achieve $6.1\%$ higher accuracy on night-time images at the $0.25\text{m}$ / $2^\circ$ threshold and 2.2\% average improvement across all evaluation thresholds (detailed results in Table~\ref{tab:aachen}). This performance narrows the gap between learning-based and traditional structure-based methods while preserving the memory advantages of coordinate regression approaches.}
    \label{fig:perf_overview}
        \vspace{-4mm}

\end{figure}
\section{Introduction}
\label{sec:intro}
Visual localization, the task of determining a camera's position and orientation from images alone, represents a core challenge in computer vision with critical applications in augmented reality, autonomous navigation, and robotics. While traditional structure-based approaches achieve robust performance through extensive feature matching and geometric verification, they impose significant computational and memory burdens that limit their practical deployment~\cite{hloc,active_search}.

Scene coordinate regression (SCR) methods~\cite{brachmann2023accelerated,GLACE2024CVPR,brachmann2018learning, focustune, DBLP:conf/cvpr/ShottonGZICF13, DBLP:conf/cvpr/ValentinNSFIT15} have emerged as an attractive alternative, excelling in small-scale environments in both accuracy and memory efficiency. However, they face perceptual aliasing as a fundamental challenge in larger environments. In such environments, visually similar landmarks, such as repeated architectural elements, often generate nearly identical local descriptors. This similarity creates ambiguity that makes it difficult to associate descriptors with their correct 3D locations and severely hampers the training effectiveness.

To address the challenge of scene ambiguity, recent works~\cite{GLACE2024CVPR, rscore} have leveraged global context to disambiguate similar local features by incorporating global descriptors. R-Score~\cite{rscore} extracts geometric relationships from covisibility graphs via the node2vec optimizer~\cite{grover2016node2vec}. In these graphs, edges represent spatial proximity between images, with images in close proximity in the covisibility graph sharing similar embeddings. These embeddings, when concatenated with local descriptors, enable effective disambiguation across scenes and substantially boost the performance of learning-based SCR methods.

However, R-Score derives global descriptors from geometric cues alone, decoupling them from visual content. This separation creates two fundamental problems: First, rich visual information that could enhance descriptor discriminability is ignored. Second, the system is vulnerable to the noise in the underlying covisibility graph, which is computed from potentially unreliable overlap scores~\cite{rau2020predicting}. When incorrect connections between dissimilar image pairs are formed, the resulting graph embeddings become partially misleading, degrading both descriptor quality and training stability.

We address these limitations by introducing a neural aggregator module that enforces geometrically-consistent global descriptors. Our key insight is that robust global descriptors should satisfy dual consistency: images should receive similar embeddings only when they are both visually similar and spatially connected in the covisibility graph. This tighter constraint mitigates the impact of erroneous connections due to noisy overlap scores while preserving meaningful geometric relationships, resulting in more discriminative and robust global descriptors.

Our aggregator learns these geometrically-consistent global descriptors through a training strategy that combines an efficient batch mining strategy relying solely on the overlap scores, and thus eliminating the need for external place labels, with a modified Generalized Contrastive Loss~\cite{leyva2023gcl} that uses a smoother distance function.

Experimental validation on challenging benchmarks demonstrates the effectiveness of our approach. As shown in Figure~\ref{fig:perf_overview}, our method achieves a $6.1\%$ improvement at the $0.25\text{m}$ / $2^\circ$ threshold compared to R-Score~\cite{rscore}, with only a 6 MB increase in memory. This gain further narrows the gap to traditional structure-based methods while preserving the memory advantages that make SCR models attractive for practical deployment.

Our contributions are as follows:
\begin{enumerate}[itemsep=0pt, parsep=0pt, topsep=0pt]    \item We introduce a neural aggregator module that learns geometrically-consistent descriptors under dual consistency constraints. Our approach produces more discriminative global descriptors that improve SCR accuracy while mitigating noisy geometric connections from the covisibility graph.
    \item We propose an effective training scheme for our aggregator module using a modified Generalized Contrastive Loss. The training scheme enables the covisibility graph to disambiguate features efficiently across the scene.
    \item We demonstrate consistent improvements across challenging benchmarks: compared to R-Score~\cite{rscore}, our method achieves average accuracy gains of $2.2\%$ and $2.5\%$ on the Aachen Day/Night~\cite{sattler2018benchmarking} and the Hyundai Department Store datasets~\cite{naverdatasets}, respectively, while maintaining similar computational and memory efficiency.
\end{enumerate}
\section{Related work}
\label{sec:rw}
We present an overview of related work in visual localization. Section~\ref{sec:rw_struct} reviews structure-based methods, which achieve high accuracy but require substantial memory resources. Section~\ref{sec:rw_learning} discusses learning-based approaches, focusing on scene coordinate regression models that predict 3D coordinates from images, which is the main focus of this paper. Section~\ref{sec:rw_local} examines the role of local descriptors in supporting these models, while Section~\ref{sec:rw_global} analyzes global descriptors that provide contextual information to enhance localization accuracy. Finally, Section~\ref{sec:rw_foundation} covers the emerging influence of foundation models in this domain.

\subsection{Structure-based visual localization} \label{sec:rw_struct}
Structure-based methods establish visual localization by matching query images against pre-built 3D scene representations. Early approaches~\cite{DBLP:conf/eccv/LiSH10, active_search, sattler2011fast} 
directly match 2D features with 3D point clouds to establish 2D-3D correspondences. 
These approaches typically construct a descriptor codebook for each 3D point by averaging descriptors from all database pixels in which the point appears. Direct matching methods can benefit from sophisticated search strategies based on both 2D-to-3D and 3D-to-2D search for additional matches~\cite{active_search, sattler2011fast}. However, they remain vulnerable to perceptual aliasing, where visually similar but spatially distinct 3D points produce incorrect correspondences.

Hierarchical approaches~\cite{peng2021megloc, hloc} address this limitation by leveraging global descriptors to streamline and refine the matching process. These approaches first retrieve database images similar to the query image using global descriptors, and then establish 2D-2D feature correspondences between retrieved and query images to obtain 2D-3D correspondences. This two-stage process significantly improves accuracy compared to direct matching.

Despite their strong performance, both direct and hierarchical approaches impose substantial memory requirements. They must maintain either complete database image descriptors~\cite{hloc,peng2021megloc} or the descriptors for 3D points~\cite{DBLP:conf/eccv/LiSH10, active_search, sattler2011fast}. Although recent advances~\cite{wang2024mad, laskar2024differentiable} have explored descriptor compression, memory requirements of structure-based algorithms still scale linearly with environment size, unlike learning-based counterparts. This motivates our focus on learning-based alternatives that offer more favorable memory characteristics.

\subsection{Learning-based visual localization} \label{sec:rw_learning}
Scene coordinate regression (SCR) models represent a fundamentally different approach to visual localization, directly predicting 3D scene coordinates for pixels in query images~\cite{DBLP:conf/cvpr/ShottonGZICF13, DBLP:conf/cvpr/ValentinNSFIT15, DBLP:journals/pami/BrachmannR22, cavallari2017fly, li2020hierarchical}. Early SCR approaches employed random forests~\cite{DBLP:conf/cvpr/ShottonGZICF13, DBLP:conf/cvpr/ValentinNSFIT15}, while subsequent approaches have adopted convolutional neural networks~\cite{DBLP:journals/pami/BrachmannR22, brachmann2016uncertainty, brachmann2018learning, brachmann2019expert, do2022learning, huang2021vs} and fully-connected architectures~\cite{brachmann2023accelerated,focustune} to improve the prediction accuracy and computational efficiency.

However, SCR approaches have historically underperformed in large-scale environments due to perceptual aliasing and ambiguous visual patterns. This fundamental challenge has motivated recent efforts to incorporate global context into SCR architectures. Recent works~\cite{GLACE2024CVPR, rscore} integrate global descriptors to provide contextual information for disambiguation. GLACE~\cite{GLACE2024CVPR} incorporates off-the-shelf descriptors with noise augmentation during training, but achieves limited robustness in large scenes due to insufficient consistency between descriptors of the same location. \mbox{R-Score}~\cite{rscore} improves scalability by learning global descriptors from a covisibility graph using node2vec embeddings~\cite{grover2016node2vec}. While R-Score demonstrated significant improvements, its graph-based representation operates independently of visual content in the image space, potentially missing important visual cues that could enhance descriptor discriminability. 

Building on these insights, we introduce a neural aggregator module that learns global descriptors informed by both the covisibility graph and the image space. This dual-informed approach enforces geometric consistency while integrating visual cues, improving robustness in large-scale environments containing significant perceptual aliasing.

\subsection{Local descriptors} \label{sec:rw_local}
Local feature descriptors form the foundation of many visual localization systems, enabling the identification and description of consistent pixels under variations in lighting, viewpoint, and scale. Classical methods~\cite{DBLP:journals/ijcv/Lowe04, DBLP:conf/eccv/BayTG06} detect invariant keypoints that can be reliably matched across different views. These approaches achieve strong real-world performance with computational efficiency, making them popular choices for structure-based localization systems~\cite{DBLP:conf/eccv/LiSH10, active_search, DBLP:conf/cvpr/IrscharaZFB09, sattler2016efficient}. 

More recent work has shifted toward learning-based local descriptors~\cite{detone2018superpoint, noh2017large, dusmanu2019d2, edstedt2024dedode}. Noh \etal~\cite{noh2017large} introduces an attention mechanism to highlight semantically meaningful local features while also estimating their confidence. \mbox{SuperPoint~\cite{detone2018superpoint}} learns keypoint detection and description on a synthetic dataset containing simple geometric shapes. 
D2-Net~\cite{dusmanu2019d2} employs a unified convolutional neural network for both dense feature description and keypoint detection, postponing the detection process to produce more stable keypoints compared to traditional methods that rely on early detection of low-level image structures.
R2D2~\cite{revaud2019r2d2} jointly learns keypoint detection, descriptor extraction, and discriminativeness prediction. This integrated approach reduces the impact of ambiguous regions, leading to more robust and reliable keypoints and descriptors.
DeDoDe~\cite{edstedt2024dedode} trains a detector using tracks from large-scale structure-from-motion and learns descriptors by optimizing a mutual nearest neighbor objective over keypoints. DeDoDe achieves strong performance in feature matching and has proven particularly effective for scene coordinate regression models~\cite{rscore}.

Our work builds on these advances by demonstrating how high-quality local descriptors can be effectively combined with learned global descriptors through our aggregator module, achieving substantial performance gains with minimal memory overhead.

\subsection{Global descriptors} \label{sec:rw_global}
Global descriptors enable finding the most similar database images to an input query image by encoding image-wide visual characteristics into compact representations~\cite{masone2021survey, DBLP:journals/trob/LowryS0LCCM16}. Current systems often reduce this problem to a similarity search in a $d$-dimensional descriptor space~\cite{hausler2021patch, DBLP:conf/cvpr/ArandjelovicGTP16, ali2023mixvpr, berton2023eigenplaces, DBLP:conf/mir/MohedanoMOSMN16, tolias2013aggregate,Izquierdo_CVPR_2024_SALAD,izquierdo2024close,leyva2023gcl}.

Global descriptors can be obtained by aggregating either local descriptors~\cite{DBLP:conf/mir/MohedanoMOSMN16, tolias2013aggregate}, multiple convolutional neural network layers~\cite{DBLP:conf/cvpr/ArandjelovicGTP16, ali2023mixvpr, berton2023eigenplaces}, or DINO features~\cite{dino, dinov2} via optimal transport~\cite{Izquierdo_CVPR_2024_SALAD}, into a single global descriptor vector. In the context of scene coordinate regression, global descriptors have emerged as crucial components providing contextual information for disambiguating local descriptors in the presence of perceptual aliasing~\cite{GLACE2024CVPR, rscore}. Our work advances this trend by introducing a dedicated aggregator module that produces high-quality global descriptors through joint optimization with both geometric and visual constraints, leading to further improvements in scene coordinate regression performance.

\subsection{Foundation models} \label{sec:rw_foundation}
Foundation models~\cite{wang2024dust3r,wang2025vggt,mast3r,reloc3r} have demonstrated strong capabilities in estimating various 3D properties, such as point clouds, depth maps, and camera poses, from just a few images, thanks to their high-quality representations. Dust3r~\cite{wang2024dust3r} pioneered this line of work by introducing pointmap regression using powerful pretrained features~\cite{weinzaepfel2023croco}. Building on this trend, Wang \etal~\cite{wang2025vggt} proposed the Visual Geometry Grounded Transformer (VGGT), a feed-forward network capable of reconstructing scenes from hundreds of views. VGGT outputs a full set of 3D attributes, including camera poses, depth maps, point maps, and 3D point tracks. While foundation models show great promise for 3D vision tasks, they are not yet specialized for visual localization, and have yet to show competitive performance with dedicated methods on large-scale benchmarks.

\section{Problem statement}
Given a training dataset $\mathcal{D} = \{\image_1, \image_2, \ldots, \image_N\}$  of $N$ images with associated ground-truth 6-DoF camera poses, our goal is to estimate the camera pose $\extrinsic_i \in \text{SE}(3)$ for a given query image $\image_i$.

We first obtain the global descriptor $\gdesc_i \in \mathbb{R}^{\dimg}$ and local descriptors $\ldesc_{ij} \in \mathbb{R}^{\diml}$ for each image $\image_i$ and its $j$-th keypoint, respectively. The local descriptors $\ldesc_{ij}$ are obtained using a pretrained off-the-shelf model~\cite{edstedt2024dedode}. The global descriptor $\gdesc_i$ is a compact representation of the entire image, obtained via the aggregator module $\agg$.

We learn a function $\Tilde{f}$ such that $\Tilde{f}(\gdesc_i, \ldesc_{ij}) = \crd_{ij}$, where $\crd_{ij} \in \mathbb{R}^{3}$ is the scene coordinate for the $j$-th local descriptor $\ldesc_{ij}$ concatenated with the global descriptor $\gdesc_{i}$ of the image $\image_i$. Therefore, $\Tilde{f}: \mathbb{R}^{\diml+\dimg} \rightarrow \mathbb{R}^{3}$ represents a mapping from descriptors to 3D coordinates. The network $\Tilde{f}$ is trained on images $\image_i \in \mathcal{D}$ and the corresponding ground-truth poses $\extrinsic_i$ by minimizing a reprojection objective given by:

\begin{equation}
\label{eq:first obj}
    \mathcal{L}_{r}(\crd_{ij}, \extrinsic_i) = ||\hat{\mathbf{y}}_{ij} - K_i\extrinsic^{-1}_i\hat{\crd}_{ij}||_2,
\end{equation}
where $K_i \in \mathbb{R}^{3\times 3}$ is the camera calibration matrix, $\pos_{ij} \in \mathbb{R}^{2}$ denotes the $j$-th keypoint coordinate in $\image_i$ and  $\crd_{ij} \in \mathbb{R}^{3}$ represents the $j$-th predicted scene coordinate. The hat \mbox{operator $\hat{\cdot}$} denotes the homogeneous coordinate representation and \mbox{$||\cdot||_2$} denotes the L2-norm. Although the optimization objective remains consistent with Eq.~\ref{eq:first obj}, more sophisticated objectives are typically used in practice to ensure stable training. For further details, we refer the reader to~\cite{brachmann2023accelerated, rscore}.

Finally, the camera pose can be computed using pairs of 2D-3D correspondences $\{(\pixel_{ij},\crd_{ij})\}$ using an off-the-shelf Perspective-n-Point solver~\cite{PoseLib, persson2018lambda}.

\section{Methodology}

\begin{figure}[t]
    \centering
    \includegraphics[width=\linewidth]{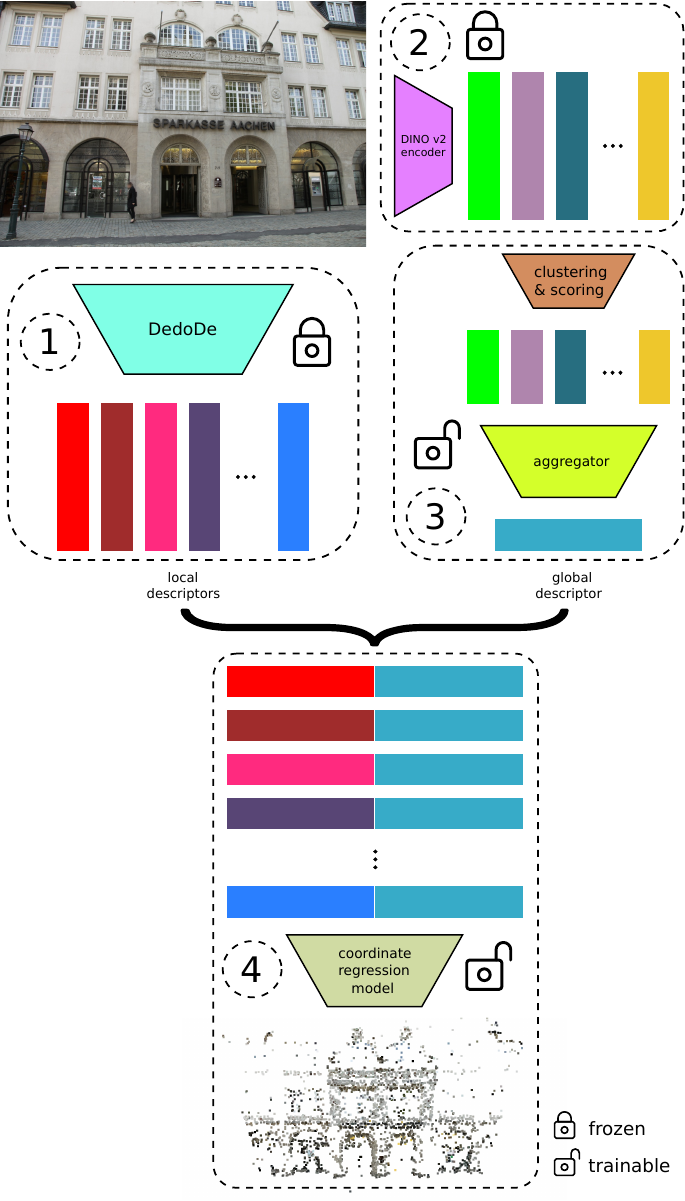}
    \caption{\textbf{System overview.} Our pipeline consists of four main components: (1) Local descriptor extraction using DeDoDe~\cite{edstedt2024dedode}, (2) DINO~\cite{dino} feature extraction for visual representation, (3) Our proposed aggregator module that learns geometrically-consistent global descriptors from DINO features using dual consistency constraints, and (4) Scene coordinate regression model that predicts 3D coordinates from concatenated local-global descriptors.}

    \label{fig:sys}
    \vspace{-4mm}
\end{figure}

Figure~\ref{fig:sys} illustrates our system architecture, which consists of four main components that work together to produce accurate camera pose estimates. The first two components extract complementary representations from input images: sparse local descriptors that capture fine-grained details, and dense visual features that encode broader visual context. The third component, our main contribution, aggregates these visual features into compact global descriptors that satisfy dual consistency constraints. The fourth component performs coordinate regression using the combined local and global descriptors.

\label{sec:method}

\textbf{Component 1: Local feature extraction.} We extract sparse local descriptors using DeDoDe~\cite{edstedt2024dedode}, which has demonstrated strong performance for coordinate regression tasks~\cite{rscore}. DeDoDe processes grayscale images to detect keypoints and outputs corresponding 256-dimensional descriptors. To reduce memory requirements and computational cost, we apply PCA~\cite{pca} compression to reduce descriptor dimensionality from 256 to 128 dimensions. Following~\cite{rscore}, we employ the L-upright weights for keypoint detection and B-upright weights for descriptor extraction.

\textbf{Component 2: Visual feature extraction.} In parallel with local descriptor extraction, we extract dense visual features using a pretrained DINO encoder~\cite{dino, dinov2}. DINO features provide rich visual representations that capture semantic content across the entire image, complementing the sparse local information of local descriptors. The DINO encoder can be fine-tuned, though we keep them frozen in our experiments.

\textbf{Component 3: Aggregator module.} Our core contribution is the aggregator module that learns geometrically-consistent global descriptors from dense DINO features. Unlike existing approaches that derive global descriptors from geometrical cues alone, our aggregator enforces dual consistency by requiring both visual similarity and geometrical constraints from the covisibility graph before assigning similar global descriptors to image pairs. The aggregator follows similar architectural principles to SALAD~\cite{Izquierdo_CVPR_2024_SALAD} but incorporates our novel training strategy for geometrical consistency. The module first projects DINO descriptors into a lower-dimensional space and computes relevance scores for each descriptor with a scoring layer. These projected descriptors are then combined through a weighted aggregation to produce a compact descriptor $\gdesc_i$. This descriptor is further compressed to $256$ dimensions using a PCA layer~\cite{pca}. The final product is concatenated with each local descriptor and passed to the fourth component, which predicts a 3D coordinate for each concatenated descriptor.

We use a modified version of the generalized contrastive loss~\cite{leyva2023gcl} for a given pair of global descriptors $\gdesc_i$ and $\gdesc_j$: 
\begin{equation}
\label{eq:gcl}
  \begin{split} 
 \mathcal{L}_{mGCL}(\gdesc_i,\gdesc_j)= \psi_{i,j}\cdot (1-\frac{\gdesc_i \cdot \gdesc_j}{\|\gdesc_i\| \|\gdesc_j\|})^2 + \\ (1-\psi_{i,j}) \cdot \max(\tau+\frac{\gdesc_i \cdot \gdesc_j}{\|\gdesc_i\| \|\gdesc_j\|},0)^2,
\end{split}
\end{equation}
where $\psi_{i,j}$ is the overlap score between the poses $\extrinsic_i$ and $\extrinsic_j$, and $\tau$ is a margin hyperparameter.

Generally, batch composition is critical for training good global descriptors~\cite{izquierdo2024close,leyva2023gcl}. We divide our batch into three groups, each of size $b$ and sample $b$ positive pairs with overlap scores $\psi_{i,j} > 0.5$, $b$ soft negative pairs with $0.25 \leq \psi_{i,j} \leq 0.5$, and $b$ random pairs with $\psi_{i,j}=0$.

\textbf{Component 4: Coordinate prediction.} The final component performs scene coordinate regression using the fused local-global descriptors. We adopt the coordinate regression architecture from R-Score~\cite{rscore}, which consists of multiple fully connected layers that map concatenated descriptors to 3D scene coordinates. These coordinates finally allow computing the camera pose $\extrinsic_i$ for the input image $\image_i$ using a Perspective-n-Point solver~\cite{PoseLib, persson2018lambda}. 

\begin{table*}[t]
    \centering
    \resizebox{\textwidth}{!}{%
    \setlength{\tabcolsep}{3pt}
    
    \begin{tabular}{clccccccccc}
    \toprule
    \multicolumn{1}{l}{} &
      &
    
    &\multicolumn{3}{c}{Aachen day} & \multicolumn{3}{c}{Aachen night}\\
    \cmidrule(l){4-6} 
    \cmidrule(l){7-9} 
    \multicolumn{1}{l}{} &  
    & \multirow{-2}{*}{\begin{tabular}[c]{@{}c@{}}Memory \\ requirement $\downarrow$ \end{tabular}}  
    & 0.25m/2$^\circ$ & 0.5m/5$^\circ$ & 5m/10$^\circ$ 
    & 0.25m/2$^\circ$ & 0.5m/5$^\circ$ & 5m/10$^\circ$ 
    & \multirow{-2}{*}{\begin{tabular}[c]{@{}c@{}}Average $\uparrow$ \\ (\%)\end{tabular}}
    
 \\
    
    \midrule
    \multirow{7}{*}{\multirow{-2}{*}{\begin{tabular}[c]{@{}c@{}}Structure-based\\ methods\end{tabular}}} 
      &hLoc (SP+SG) \cite{hloc, sarlin2020superglue}&7.82 GB&\textbf{89.6}&\textbf{95.4}&\textbf{98.8}&{\ul86.7}&\textbf{93.9}&\textbf{100.0}&\textbf{94.1}\\    
      &DeViLoc \cite{deviloc}& $\approx$7.82 GB&{\ul87.4}&{\ul94.8}&{\ul98.2}&\textbf{87.8}&\textbf{93.9}&\textbf{100.0}&93.7\\    
      &AS (SIFT) \cite{active_search}&750 MB  &85.3&92.2&97.9&39.8&49.0&64.3&71.4\\
      &Cascaded  \cite{cheng2019cascaded}&140 MB  &76.7&88.6&95.8&33.7&48.0&62.2&67.5\\
      &Squeezer \cite{yang2022scenesqueezer}&240 MB  &75.5&89.7&96.2&50.0&67.3&78.6&76.2\\
      &PixLoc \cite{sarlin21pixloc}&2.13 GB  &64.3&69.3&77.4&51.1&55.1&67.3&64.1\\
    \midrule

    \multirow{8}{*}{\multirow{-2}{*}{\begin{tabular}[c]{@{}c@{}}Learning-based\\ methods\end{tabular}}} 

    &ACE ($\times50$) \cite{brachmann2023accelerated}&205 MB   &6.9&17.2&50.0&0.0&1.0&5.1&13.4\\
          &GLACE \cite{GLACE2024CVPR}&27 MB &8.6 & 20.8 & 64.0  &1.0&1.0&17.3&18.8\\

    &ESAC ($\times50$) \cite{brachmann2019expert}&1.31 GB   &42.6&59.6&75.5&6.1&10.2&18.4&35.4\\
    &HSCNet++ \cite{wang2024hscnet++}&274 MB   &72.7&81.6&91.4&43.9&57.1&76.5&70.5\\
    &Neumap \cite{tang2023neumap}&1.26 GB   &\textbf{80.8}&\textbf{90.9}&{95.6}&{48.0}&{67.3}&{87.8}&{78.4}\\
      &R-Score \cite{rscore}&47 MB   &79.0&88.5&{\ul96.4}&{\ul66.3}&{\ul89.8}&{\ul96.9}&{\ul86.1}\\

     &Ours &53 MB& {\ul80.1}&{\ul89.8}&\textbf{97.0}&\textbf{72.4}&\textbf{91.8}&\textbf{99.0}&\textbf{88.3}\\
    
    \bottomrule
    \end{tabular}
    }
    \caption{\textbf{Aachen day/night Dataset \cite{sattler2018benchmarking}.}
    We report the percentage of query images successfully localized under varying error thresholds. Under the $0.25\text{m}$ / $2^\circ$ threshold, our system outperforms R-Score~\cite{rscore} by $1.1\%$ and $6.1\%$ under day and night conditions, respectively. We maintain a memory footprint of 53 MB, which is only 6 MB higher than that of R-Score~\cite{rscore}. Best and second-best results for each class are highlighted in \textbf{bold} and {\ul underlined}. We report results using the strongest variant of R-Score~\cite{rscore} (DeDoDe~\cite{edstedt2024dedode} with depth supervision).
    }
    
         \label{tab:aachen}
    \vspace{-1mm}

    \end{table*}
\begin{table*}[t]
    \centering
   \resizebox{\textwidth}{!}{%
    \setlength{\tabcolsep}{3pt}
    
    \begin{tabular}{lcccccccccccc}
    \toprule
    \multicolumn{1}{l}{} &
      &
    
    \multicolumn{3}{c}{Dept. 1F} & \multicolumn{3}{c}{Dept. 4F} & \multicolumn{3}{c}{Dept. B1}\\
    \cmidrule(l){3-5} 
    \cmidrule(l){6-8} 
    \cmidrule(l){9-11} 
    
    & \multirow{-2}{*}{\begin{tabular}[c]{@{}c@{}}Memory \\ requirement $\downarrow$ \end{tabular}}  
    & 0.1m/1$^\circ$ & 0.25m/2$^\circ$ & 1m/5$^\circ$ 
    & 0.1m/1$^\circ$ & 0.25m/2$^\circ$ & 1m/5$^\circ$ 
    & 0.1m/1$^\circ$ & 0.25m/2$^\circ$ & 1m/5$^\circ$ 
        & \multirow{-2}{*}{\begin{tabular}[c]{@{}c@{}}Average $\uparrow$ \\ (\%)\end{tabular}}

 \\
    
    \midrule
      hLoc (R2D2) \cite{hloc,revaud2019r2d2}&150 GB&{\ul80.6}&{\ul84.3}&{\ul89.4}&{\ul85.3}&{\ul91.0}&{\ul93.1}&{\ul75.2}&{\ul80.3}&{\ul87.6}&{\ul85.2}\\    
      hLoc (D2-net) \cite{hloc,dusmanu2019d2}&362 GB&78.0&82.8&88.0&84.2&89.8&92.0&73.7&79.3&87.2&83.9\\    
      DeViLoc \cite{deviloc}&$\approx$362 GB& \textbf{86.9}&\textbf{91.5}&\textbf{96.3}&\textbf{88.7}&\textbf{93.7}&\textbf{96.1}&\textbf{78.5}&\textbf{84.2}&\textbf{93.7}&\textbf{90.0}\\    
    \midrule
    ACE ($\times50$) \cite{brachmann2023accelerated}&205 MB   &14.1 & 54.4 & 75.5  &27.3&70.9&84.1&2.7&14.4&29.3&41.4\\
   ESAC ($\times50$) \cite{brachmann2019expert}&1.4 GB   &43.3&66.3&77.0&45.2&62.5&73.1&3.5&8.2&12.6&43.5\\
      GLACE \cite{GLACE2024CVPR}&42 MB &5.6 & 21.3 & 48.6  &8.4&29.8&51.6&0.9&4.4&11.9&20.3\\
R-Score (LoFTR) \cite{rscore, sun2021loftr}&102 MB  &{67.3}&{84.5}&\textbf{92.6}&
{70.5}&{87.0}&{\ul92.9}&{30.8}&{53.7}&{72.7}&{72.4}\\
      R-Score (DeDoDe) \cite{rscore, edstedt2024dedode}&102 MB   &63.9&83.3&90.8&76.7&89.3&\textbf{93.0}&61.5&77.6&88.8&80.5\\

     Ours &109 MB& \textbf{70.4}&\textbf{85.6}&{\ul92.0}&\textbf{77.6}&{\ul88.8}&{92.1}&\textbf{66.6}&\textbf{81.7}&\textbf{92.6}&\textbf{83.0}\\

    \bottomrule
    \end{tabular}
   }
    \caption{\textbf{Hyundai Department Store Dataset \cite{naverdatasets}.} 
    We report the percentage of query images successfully localized under different thresholds, along with average memory requirements across three scenes. Our system is $2.5\%$ more accurate than R-Score~\cite{rscore}, while consuming only 7 MB extra on average. We report results using the strongest variants of R-Score~\cite{rscore} with depth supervision. Best and second-best results for each class are highlighted in \textbf{bold} and {\ul underlined}.}
         \label{tab:hyundai}

    \end{table*}

\section{Implementation details}
\label{subsec:impl}

This section details our dual-consistent SCR model, covering graph construction, aggregator architecture and training procedures, and coordinate regression settings.

\textbf{Covisibility graph.}
Following R-Score~\cite{rscore}, we construct the covisibility graph using pose-based overlap estimation~\cite{rau2020predicting}. For each training image, we sample random pixel coordinates and unproject them using uniformly sampled depth values to generate 3D point hypotheses. These 3D points are then projected onto the other camera view to compute pairwise overlap scores based on the fraction of points that fall within the image boundaries. An edge is added between two images to the covisibility graph if their overlap score exceeds a threshold of 0.2.

\textbf{Aggregator module.} Our aggregator module processes dense DINO features to produce compact global descriptors. We employ the base version of DINOv2 (ViT-B/14)~\cite{dinov2}, which generates 768-dimensional features. The aggregator consists of two main layers. The cluster layer reduces the dimensionality of DINO features from 768 to 128 dimensions using a trainable linear projection. The scoring layer then computes attention weights for each spatial location, enabling the aggregator to focus on the most relevant visual regions. The attention mechanism produces a weighted combination of the compressed features, resulting in an intermediate descriptor of 2304 dimensions. 
The intermediate descriptor is compressed to $256$ dimensions using a PCA layer~\cite{pca}. See Section~2 of the Supplementary Material for an ablation study on different dimensions.

\textbf{Aggregator training.} We train the aggregator module for 10,000 iterations using a batch size of 64. The optimization employs AdamW~\cite{adam_kingma} with scene-specific learning rates: $3 \times 10^{-3}$ for outdoor scenes and $1 \times 10^{-3}$ for indoor scenes. The margin hyperparameter $\tau$ in Eq.~\ref{eq:gcl} is set to $0.5$. The training takes less than an hour using a single NVIDIA H100 GPU. We train the aggregator without using any data augmentation, as the DINO features~\cite{dino} are already robust to various transformations.

\textbf{Coordinate regression settings.} We follow the same configuration as R-Score~\cite{rscore}. Specifically, we use the DeDoDe keypoint encoder~\cite{edstedt2024dedode}, which produces 128-dimensional local descriptors after being compressed using a PCA model.
During testing, we adopt the multi-hypothesis strategy from R-Score~\cite{rscore}, retrieving the top-$10$ hypotheses per query image. For this purpose, we use the SALAD~\cite{Izquierdo_CVPR_2024_SALAD} global descriptors with product quantization. We refer the reader to Table~\ref{tab:ab1} for different results with other state-of-the-art global descriptors~\cite{ali2024boq,ali2023mixvpr,berton2025megaloc} and Section~5 of the Supplementary Material for other local descriptors.

\textbf{Coordinate regression training.} We adopt the same optimization scheme as R-Score~\cite{rscore} to train the encoder module using a buffer of $128 \times 10^6$ examples and a batch size of $320 \times 10^3$. Instead of randomly sampling keypoints to populate the training buffer, we use heuristic sampling as described in FocusTune~\cite{focustune} using the SfM model if available. The training takes approximately 11 hours on average using a single NVIDIA H100 GPU. Finally, we use depth supervision and graph-based augmentation~\cite{rscore} for all of our experiments.

\section{Evaluation}
This section presents a comprehensive evaluation of our system.
We report the performance gains in both outdoor and indoor scenarios 
in Section~\ref{sec:eval_quant}. We also present a qualitative evaluation of our method in Section~\ref{sec:eval_qual}. Finally, Section~\ref{sec:ablation} provides ablation studies to assess the impact of key components in our system.

\label{sec:eval}


\begin{figure*}[t]
    \centering
    \includegraphics[width=\textwidth]{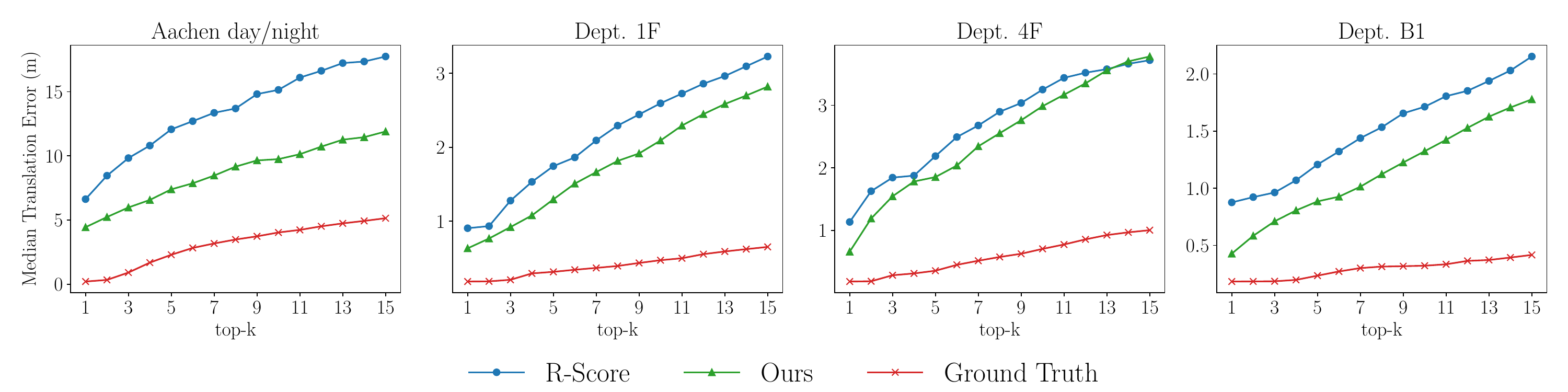}
    \caption{\textbf{Median translation error.} We plot the median translation error of the top-$k$ retrievals for all the training images. The results show that our learned global descriptors (in green) offer more relevant retrievals over the node2vec graph embeddings (in blue) of R-Score~\cite{rscore}. We also plot the ground-truth errors (in red) for reference.}
    \label{fig:top_k_errors}
        \vspace{-2mm}

\end{figure*}

\begin{figure*}[t]
    \centering
    \includegraphics[width=\textwidth]{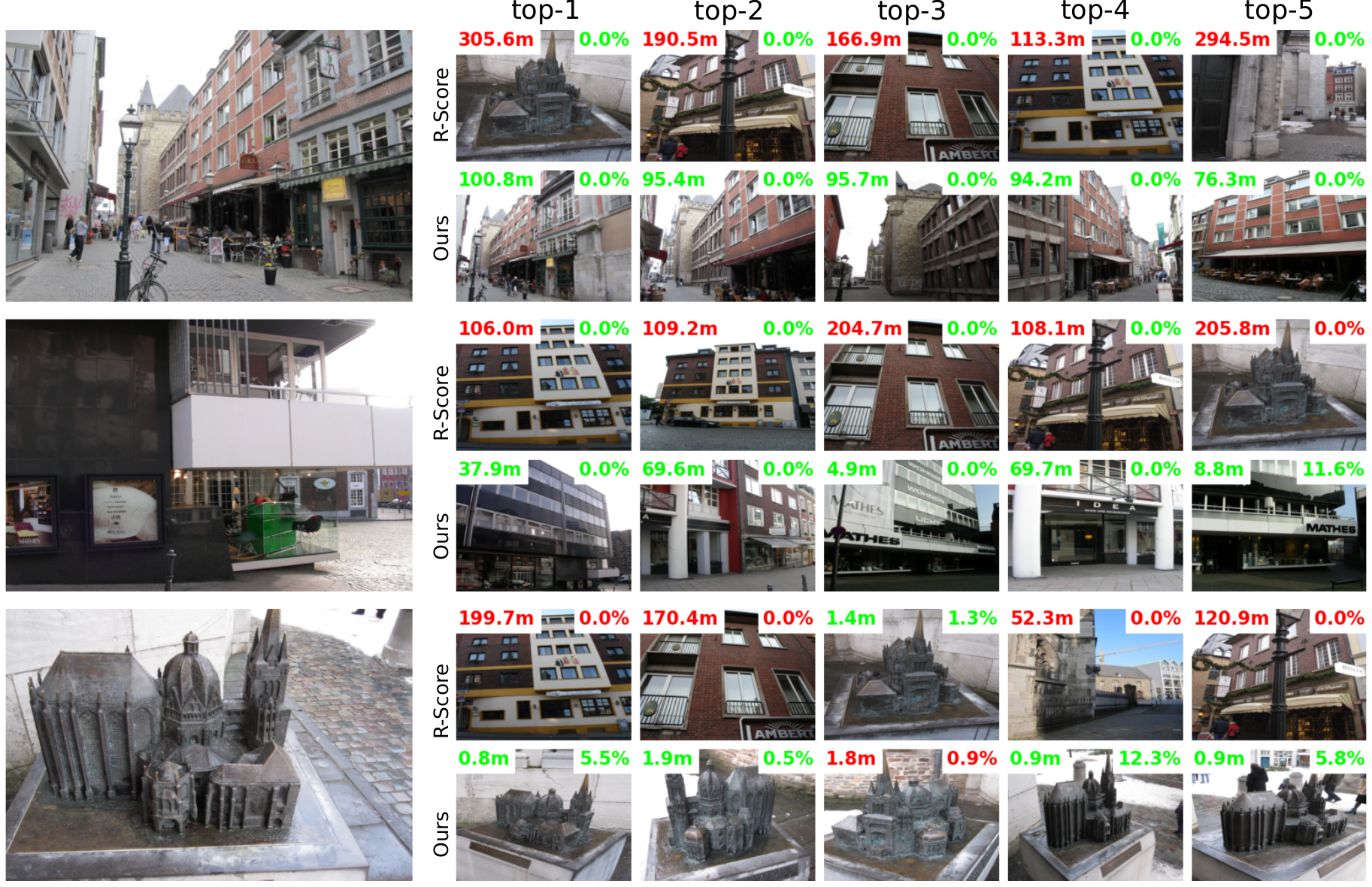}
    \caption{\textbf{Qualitative results.} We visualize the top-$5$ retrievals for three images from the training set of the Aachen Day/Night dataset~\cite{sattler2018benchmarking}. For each query (left-most column), we show results using graph embeddings from R-Score~\cite{rscore} (\textit{top}) and our learned global descriptors (\textit{bottom}). On top of each retrieval, we plot the translation error and the overlap score between the retrievals and the query. Under heavy noise in the covisibility graph (evidenced by low overlap scores), the graph embeddings retrieve nearby but not exact structures, while our global descriptors retrieve nearby and relevant ones. }
    \label{fig:qualitative}
        \vspace{-4mm}

\end{figure*}




\subsection{Quantitative evaluation}
\label{sec:eval_quant}
We evaluate our method on the Aachen day/night dataset~\cite{sattler2018benchmarking} and the Hyundai Department Store dataset~\cite{naverdatasets}. 

\textbf{Aachen day/night dataset~\cite{sattler2018benchmarking}.} Table~\ref{tab:aachen} shows that our method enhances the accuracy of scene coordinate regression models~\cite{rscore,GLACE2024CVPR,brachmann2023accelerated,focustune}. Our method outperforms the state-of-the-art R-Score~\cite{rscore}, achieving improvements of $1.1\%$ and $6.1\%$ for the $0.25\text{m}$ and $2^\circ$ threshold on daytime and nighttime images, respectively, and yielding an average gain of $2.2\%$ across all thresholds. 
Our method continues to close the gap to structure-based methods, such as hLoc~\cite{hloc} and DeViLoc~\cite{deviloc}, achieving only $5.8\%$ and $5.4\%$ lower localization accuracy, respectively, while using just $0.67\%$ of hLoc's memory.

\textbf{Hyundai Department Store dataset~\cite{naverdatasets}.} Table~\ref{tab:hyundai} shows that our method consistently outperforms R-Score~\cite{rscore} by $6.5\%$, $0.9\%$, and $4.1\%$ for floors 1F, 4F, and B1 respectively, for the $0.1\text{m}$ and $1^\circ$ threshold. Overall, we achieve a $2.5\%$ improvement in average localization accuracy while requiring only 7 MB additional memory. Compared to hLoc~\cite{hloc}, our method further closes the gap to just $2.2\%$, using only $0.07\%$ of hLoc’s memory.

\textbf{Median translational error.} We plot the median translation error of top-$k$ retrievals across all four scenes in Figure~\ref{fig:top_k_errors}. Our learned global descriptors reduce the error by around 5 meters on the Aachen dataset and 1 meter on the Hyundai Department~1F and B1 scenes. However, the performance gain is weaker on Department~4F, possibly due to the prevalence of highly similar structures, which makes accurate retrieval more challenging.

\subsection{Qualitative evaluation} 
\label{sec:eval_qual}

Figure~\ref{fig:qualitative} shows the top-$5$ retrievals for three randomly selected queries from the Aachen Day/Night training set~\cite{sattler2018benchmarking}, comparing R-Score's graph embeddings (top) with our learned global descriptors (bottom). The translation errors and the overlap scores are shown in the top-left and top-right corners, respectively. Our descriptors retain strong retrieval performance even under noisy geometrical constraints (evidenced by low overlap scores), due to their consistency with both the image space and the covisibility graph. In contrast, R-Score might retrieve irrelevant images whose visual content does not match the query image.

\subsection{Ablation studies}
\label{sec:ablation}

\textbf{Buffer sampling.} Consistent with \cite{focustune}, we observe that using the SfM model to guide buffer sampling results in a more efficient training process and improved performance. As shown in Table~\ref{tab:ab2}, integrating FocusTune~\cite{focustune} yields nearly a $10\%$ improvement on the Aachen nighttime test set, highlighting the impact of effective buffer sampling for SCR models.

\textbf{Global descriptors at query time.} At test time, multiple hypotheses can be retrieved using any standard global descriptor model~\cite{DBLP:conf/cvpr/ArandjelovicGTP16, Izquierdo_CVPR_2024_SALAD, izquierdo2024close, ali2023mixvpr}. Table~\ref{tab:ab1} presents the performance of several popular methods. Among them, SALAD~\cite{Izquierdo_CVPR_2024_SALAD} performs best on daytime queries, while BoQ~\cite{ali2024boq} excels at nighttime. Notably, even with NetVLAD~\cite{DBLP:conf/cvpr/ArandjelovicGTP16}, our system achieves significant improvements over R-Score~\cite{rscore}, gaining $1.3\%$ and $8.2\%$ at the $0.25\text{m}$ and $2^\circ$ threshold for day and night queries, respectively. These results highlight the superior quality of our learned global descriptors. We use SALAD~\cite{Izquierdo_CVPR_2024_SALAD} in all of our experiments.

\textbf{Loss function for the aggregator module.} Table~\ref{tab:ab2} presents the impact of different loss functions used in the aggregator module. Our modified GCL (mGCL) consistently outperforms the original Generalized Contrastive Loss (GCL)~\cite{leyva2023gcl}, yielding a $2.1\%$ accuracy gain on daytime images and a $3\%$ gain on nighttime images for the $0.25\text{m}$ and $2^\circ$ threshold on the Aachen dataset, highlighting the effectiveness of our mGCL in learning more discriminative global descriptors. The hyperparameter margin $\tau$ improves the accuracy of our system by $2-3\%$ compared to the default value of $0.5$ (see Section~1 of the Supplementary Material). We use mGCL with $\tau=0.5$ in all of our experiments. 

\textbf{Number of trainable blocks in the DINO encoder.} We also evaluate the impact of the number of trainable blocks in the DINO encoder (the second component in Figure~\ref{fig:sys}), which is used to extract the DINO features for the aggregator module. Table~\ref{tab:ab_layers} shows that training the last two blocks can yield 1-2\% additional accuracy, but also requires approximately 60 MB more memory. As a result, we keep the DINO encoder frozen in all of our experiments.

\begin{table}[t]
    \centering
    \setlength{\tabcolsep}{5pt}
    \scriptsize
    \begin{tabular}{l*{3}{r}*{3}{r}}
    \toprule
                   & \multicolumn{3}{c}{Aachen Day} & \multicolumn{3}{c}{Aachen Night} \\
    \midrule
        R-Score (with NetVLAD \cite{DBLP:conf/cvpr/ArandjelovicGTP16})   & 79.0 & 88.5 & 96.4 & 66.3 & 89.8 & 96.9 \\
        Ours (with NetVLAD \cite{DBLP:conf/cvpr/ArandjelovicGTP16})       & {79.9}&\textbf{91.0}&\textbf{97.2}&{73.5}&{90.8}&{96.9} \\
        Ours (with MegaLoc \cite{berton2025megaloc})          & {79.4}&{90.8}&{96.8}&{73.5}&\textbf{91.8}&\textbf{99.0} \\
        Ours (with MixVPR \cite{ali2023mixvpr})          & {79.9}&{90.8}&\textbf{97.2}&{73.5}&{89.8}&{96.9}  \\
        Ours (with EigenPlaces \cite{berton2023eigenplaces})          & {79.5}&{90.3}&{96.0}&{70.4}&\textbf{91.8}&{96.9} \\
        Ours (with BoQ \cite{ali2024boq})          & {79.7}&{90.5}&{96.8}&\textbf{74.5}&\textbf{91.8}&\textbf{99.0}\\
        Ours (with SALAD \cite{Izquierdo_CVPR_2024_SALAD})          & \textbf{80.3}&{90.3}&{97.1}&{72.4}&{90.8}&\textbf{99.0} \\
    \bottomrule
    \end{tabular}%
    \caption{\textbf{Global descriptors at test time.} To retrieve multiple hypotheses at test time, we can use any off-the-shelf global descriptors~\cite{DBLP:conf/cvpr/ArandjelovicGTP16, ali2023mixvpr, berton2023eigenplaces, ali2024boq, Izquierdo_CVPR_2024_SALAD}. We use SALAD~\cite{Izquierdo_CVPR_2024_SALAD} in all of our experiments.
    }
        \vspace{-1mm}

    \label{tab:ab1}
\end{table}
\begin{table}[t]
    \centering
    \scriptsize
    \setlength{\tabcolsep}{4pt}
    \begin{tabular}{l*{3}{r}*{3}{r}}
    \toprule
                   & \multicolumn{3}{c}{Aachen Day} & \multicolumn{3}{c}{Aachen Night} \\
    \midrule
        Ours (vanilla, NetVLAD, GCL)       &  72.3&85.8&95.4&58.2&80.6&94.9\\
        Ours (FocusTune, NetVLAD, GCL)       & {77.9}&{89.6}&{96.4}&{68.4}&{88.8}&{96.9}\\
        Ours (FocusTune, SALAD, GCL)       &  {78.2}&{89.7}&{96.4}&{69.4}&{90.8}&{99.0}\\
        Ours (FocusTune, SALAD, mGCL)          & \textbf{80.3}&\textbf{90.3}&\textbf{97.1}&\textbf{72.4}&\textbf{90.8}&\textbf{99.0} \\
    \bottomrule
    \end{tabular}%
    \caption{\textbf{Different components.} Using FocusTune~\cite{focustune}, better global descriptors, and our mGCL in Eq.~\ref{eq:gcl} all further improve the performance by $8-14 \%$ on the tightest threshold.}
    \label{tab:ab2}
        \vspace{-1mm}

\end{table}
\begin{table}[t]
    \centering
    \scriptsize
    \setlength{\tabcolsep}{5pt}
    \begin{tabular}{lcl*{3}{r}*{3}{r}}
    \toprule
                  &Memory&  \multicolumn{3}{c}{Aachen Day} & \multicolumn{3}{c}{Aachen Night} \\
    \midrule
        Ours (all frozen)       &6 MB&  {80.3}&{90.3}&{97.1}&\textbf{72.4}&{90.8}&{99.0}\\
        Ours (last two)       & 62 MB& \textbf{80.6}&\textbf{91.3}&{97.1}&{70.4}&\textbf{93.9}&{99.0}\\
        Ours (last four)          &119 MB& {79.9}&{90.9}&\textbf{97.2}&{70.4}&{91.8}&{99.0}\\
    \bottomrule
    \end{tabular}%
    \caption{\textbf{DINO encoder's trainable blocks.} Training the last two blocks can yield some extra performance, but also requires almost 60 MB extra. Therefore, to keep our system lightweight, we keep the encoder frozen in all of our experiments.}
    \label{tab:ab_layers}
        \vspace{-4mm}

\end{table}

\section{Conclusion}

\label{sec:conclusion}

We introduce a novel aggregator module that learns geometrically-consistent global descriptors by enforcing dual consistency between visual similarity and geometrical connectivity via the covisibility graph. Unlike existing approaches that rely solely on the covisibility graph~\cite{rscore} or off-the-shelf models~\cite{GLACE2024CVPR}, our method integrates both geometrical relationships and visual content to improve resilience to perceptual aliasing and noisy covisibility graphs.

Our approach delivers substantial improvements in localization accuracy while requiring minimal additional memory. A key component is our batch mining strategy that depends solely on the overlap score, removing the need for manual place labels and enabling faster convergence during training. This makes our method more accessible for real-world deployment across diverse environments.
We believe this work opens promising directions for designing more efficient scene coordinate regression systems that more tightly integrate local and global descriptors.



{\footnotesize
\noindent\textbf{Acknowledgements:} This research was partially supported by the QUT Centre for Robotics, an ARC Laureate Fellowship FL210100156 to MM, and an ARC DECRA Fellowship DE240100149 to TF.
}

{\small
\bibliographystyle{ieee_fullname}
\bibliography{egbib}

@String(PAMI = {IEEE Trans. Pattern Anal. Mach. Intell.})

@String(IJCV = {Int. J. Comput. Vis.})

@String(CVPR= {IEEE Conf. Comput. Vis. Pattern Recog.})

@String(ICCV= {Int. Conf. Comput. Vis.})

@String(ECCV= {Eur. Conf. Comput. Vis.})

@String(NIPS= {Adv. Neural Inform. Process. Syst.})

@String(ICLR = {Int. Conf. Learn. Represent.})

@String(CVPRW= {IEEE Conf. Comput. Vis. Pattern Recog. Worksh.})

@String(PAMI  = {IEEE TPAMI})

@String(IJCV  = {IJCV})

@String(CVPR  = {CVPR})

@String(ICCV  = {ICCV})

@String(ECCV  = {ECCV})

@String(NIPS  = {NeurIPS})

@String(ICLR  = {ICLR})

@String(CVPRW= {CVPRW})

@String(ICCV= {IEEE Int. Conf. Comput. Vis.})

@String(WACV = {IEEE Winter Conf. Applicat. Comput. Vis.})

@String(TRO = {IEEE Trans. Robot.})

@String(ICMR = {Int. Conf. Multimedia Retrieval})

@inproceedings{sun2021loftr,
  title={LoFTR: Detector-free local feature matching with transformers},
  author={Sun, Jiaming and Shen, Zehong and Wang, Yuang and Bao, Hujun and Zhou, Xiaowei},
  booktitle=CVPR,
  pages={8922--8931},
  year={2021}
}

@inproceedings{berton2025megaloc,
  title={Megaloc: One retrieval to place them all},
  author={Berton, Gabriele and Masone, Carlo},
  booktitle=CVPRW,
  pages={2861--2867},
  year={2025}
}

@inproceedings{ali2024boq,
  title={BoQ: A place is worth a bag of learnable queries},
  author={Ali-Bey, Amar and Chaib-draa, Brahim and Gigu{\`e}re, Philippe},
  booktitle=CVPR,
  pages={17794--17803},
  year={2024}
}

@article{dinov2,
  author       = {Maxime Oquab and
                  Timoth{\'{e}}e Darcet and
                  Th{\'{e}}o Moutakanni and
                  Huy V. Vo and
                  Marc Szafraniec and
                  Vasil Khalidov and
                  Pierre Fernandez and
                  Daniel Haziza and
                  Francisco Massa and
                  Alaaeldin El{-}Nouby and
                  Mido Assran and
                  Nicolas Ballas and
                  Wojciech Galuba and
                  Russell Howes and
                  Po{-}Yao Huang and
                  Shang{-}Wen Li and
                  Ishan Misra and
                  Michael Rabbat and
                  Vasu Sharma and
                  Gabriel Synnaeve and
                  Hu Xu and
                  Herv{\'{e}} J{\'{e}}gou and
                  Julien Mairal and
                  Patrick Labatut and
                  Armand Joulin and
                  Piotr Bojanowski},
  title        = {DINOv2: Learning Robust Visual Features without Supervision},
  journal      = {Trans. Mach. Learn. Res.},
  year         = {2024}
}

@inproceedings{laskar2024differentiable,
  title={Differentiable product quantization for memory efficient camera relocalization},
  author={Laskar, Zakaria and Melekhov, Iaroslav and Benbihi, Assia and Wang, Shuzhe and Kannala, Juho},
  booktitle=ECCV,
  pages={470--489},
  year={2024},
  organization={Springer}
}

@inproceedings{wang2024mad,
  title={{MAD-DR}: Map compression for visual localization with matchness aware descriptor dimension reduction},
  author={Wang, Qiang},
  booktitle=ECCV,
  pages={261--278},
  year={2024},
}

@inproceedings{tolias2013aggregate,
  title={To aggregate or not to aggregate: Selective match kernels for image search},
  author={Tolias, Giorgos and Avrithis, Yannis and J{\'e}gou, Herv{\'e}},
  booktitle=ICCV,
  pages={1401--1408},
  year={2013}
}

@article{pca,
  title={Finding structure with randomness: Probabilistic algorithms for constructing approximate matrix decompositions},
  author={Halko, Nathan and Martinsson, Per-Gunnar and Tropp, Joel A},
  journal={SIAM review},
  volume={53},
  number={2},
  pages={217--288},
  year={2011},
  publisher={SIAM}
}

@inproceedings{detone2018superpoint,
  title={Superpoint: Self-supervised interest point detection and description},
  author={DeTone, Daniel and Malisiewicz, Tomasz and Rabinovich, Andrew},
  booktitle=CVPRW,
  pages={224--236},
  year={2018}
}

@inproceedings{noh2017large,
  title={Large-scale image retrieval with attentive deep local features},
  author={Noh, Hyeonwoo and Araujo, Andre and Sim, Jack and Weyand, Tobias and Han, Bohyung},
  booktitle=ICCV,
  pages={3456--3465},
  year={2017}
}

@inproceedings{edstedt2024dedode,
  title={DeDoDe: Detect, don’t describe—Describe, don’t detect for local feature matching},
  author={Edstedt, Johan and B{\"o}kman, Georg and Wadenb{\"a}ck, M{\aa}rten and Felsberg, Michael},
  booktitle={Int. Conf. 3D Vision},
  pages={148--157},
  year={2024},
}

@inproceedings{izquierdo2024close,
  title={Close, But Not There: Boosting Geographic Distance Sensitivity in Visual Place Recognition},
  author={Izquierdo, Sergio and Civera, Javier},
  booktitle=ECCV,
  pages={240--257},
  year={2024},
}

@inproceedings{leyva2023gcl,
  title={Data-efficient large scale place recognition with graded similarity supervision},
  author={Leyva-Vallina, Mar{\'\i}a and Strisciuglio, Nicola and Petkov, Nicolai},
  booktitle=CVPR,
  pages={23487--23496},
  year={2023}
}

@inproceedings{dino,
  title={Emerging Properties in Self-Supervised Vision Transformers},
  author={Caron, Mathilde and Touvron, Hugo and Misra, Ishan and J\'egou, Herv\'e  and Mairal, Julien and Bojanowski, Piotr and Joulin, Armand},
  pages={9650--9660},
  booktitle=ICCV,
  year={2021}
}

@inproceedings{naverdatasets,
  title={Large-scale localization datasets in crowded indoor spaces},
  author={Lee, Donghwan and Ryu, Soohyun and Yeon, Suyong and Lee, Yonghan and Kim, Deokhwa and Han, Cheolho and Cabon, Yohann and Weinzaepfel, Philippe and Gu{\'e}rin, Nicolas and Csurka, Gabriela and others},
  booktitle=CVPR,
  pages={3227--3236},
  year={2021}
}

@inproceedings{grover2016node2vec,
  title={node2vec: Scalable feature learning for networks},
  author={Grover, Aditya and Leskovec, Jure},
  booktitle={ACM SIGKDD},
  pages={855--864},
  year={2016}
}

@article{wang2024hscnet++,
  title={Hscnet++: Hierarchical scene coordinate classification and regression for visual localization with transformer},
  author={Wang, Shuzhe and Laskar, Zakaria and Melekhov, Iaroslav and Li, Xiaotian and Zhao, Yi and Tolias, Giorgos and Kannala, Juho},
  journal=IJCV,
  volume={132},
  number={7},
  pages={2530--2550},
  year={2024},
  publisher={Springer}
}

@inproceedings{persson2018lambda,
  title={Lambda twist: An accurate fast robust perspective three point {(P3P)} solver},
  author={Persson, Mikael and Nordberg, Klas},
  booktitle=ECCV,
  pages={318--332},
  year={2018}
}

@inproceedings{GLACE2024CVPR,
      title     = {{GLACE:} Global Local Accelerated Coordinate Encoding},
      author    = {Fangjinhua Wang and Xudong Jiang and Silvano Galliani and Christoph Vogel and Marc Pollefeys},
      booktitle = CVPR,
      year      = {2024},
pages = {21562--21571}
  }

@inproceedings{sarlin2020superglue,
  title={Superglue: Learning feature matching with graph neural networks},
  author={Sarlin, Paul-Edouard and DeTone, Daniel and Malisiewicz, Tomasz and Rabinovich, Andrew},
  booktitle=CVPR,
  pages={4938--4947},
  year={2020}
}

@inproceedings{DBLP:conf/eccv/BayTG06,
  author       = {Herbert Bay and
                  Tinne Tuytelaars and
                  Luc Van Gool},
  title        = {{SURF:} Speeded Up Robust Features},
  booktitle    = ECCV,
  pages        = {404--417},
  year         = {2006}
}

@inproceedings{DBLP:conf/cvpr/ArandjelovicGTP16,
  author       = {Relja Arandjelovic and
                  Petr Gron{\'{a}}t and
                  Akihiko Torii and
                  Tom{\'{a}}s Pajdla and
                  Josef Sivic},
  title        = {NetVLAD: {CNN} Architecture for Weakly Supervised Place Recognition},
  booktitle    = CVPR,
  pages        = {5297--5307},
  year         = {2016}
}

@inproceedings{DBLP:conf/mir/MohedanoMOSMN16,
  author       = {Eva Mohedano and
                  Kevin McGuinness and
                  Noel E. O'Connor and
                  Amaia Salvador and
                  Ferran Marqu{\'{e}}s and
                  Xavier Gir{\'{o}}{-}i{-}Nieto},
  title        = {Bags of Local Convolutional Features for Scalable Instance Search},
  booktitle    = ICMR,
  pages        = {327--331},
  year         = {2016}
}

@inproceedings{cavallari2017fly,
  title={On-the-fly adaptation of regression forests for online camera relocalisation},
  author={Cavallari, Tommaso and Golodetz, Stuart and Lord, Nicholas A and Valentin, Julien and Di Stefano, Luigi and Torr, Philip HS},
  booktitle=CVPR,
  pages={4457--4466},
  year={2017}
}

@article{sattler2016efficient,
  title={Efficient \& effective prioritized matching for large-scale image-based localization},
  author={Sattler, Torsten and Leibe, Bastian and Kobbelt, Leif},
  journal=PAMI,
  volume={39},
  number={9},
  pages={1744--1756},
  year={2016},
}

@inproceedings{adam_kingma,
  author       = {Diederik P. Kingma and
                  Jimmy Ba},
  title        = {Adam: {A} Method for Stochastic Optimization},
  booktitle    = ICLR,
  year         = {2015}
}

@article{masone2021survey,
  title={A survey on deep visual place recognition},
  author={Masone, Carlo and Caputo, Barbara},
  journal={IEEE Access},
  volume={9},
  pages={19516--19547},
  year={2021},
}

@article{DBLP:journals/trob/LowryS0LCCM16,
  author       = {Stephanie M. Lowry and
                  Niko S{\"{u}}nderhauf and
                  Paul Newman and
                  John J. Leonard and
                  David D. Cox and
                  Peter I. Corke and
                  Michael J. Milford},
  title        = {Visual Place Recognition: {A} Survey},
  journal      = TRO,
  volume       = {32},
  number       = {1},
  pages        = {1--19},
  year         = {2016}
}

@inproceedings{ali2023mixvpr,
  title={Mixvpr: Feature mixing for visual place recognition},
  author={Ali-Bey, Amar and Chaib-Draa, Brahim and Giguere, Philippe},
  booktitle=WACV,
  pages={2998--3007},
  year={2023}
}

@inproceedings{berton2023eigenplaces,
  title={Eigenplaces: Training viewpoint robust models for visual place recognition},
  author={Berton, Gabriele and Trivigno, Gabriele and Caputo, Barbara and Masone, Carlo},
  booktitle=ICCV,
  pages={11080--11090},
  year={2023}
}

@inproceedings{hausler2021patch,
  title={Patch-netvlad: Multi-scale fusion of locally-global descriptors for place recognition},
  author={Hausler, Stephen and Garg, Sourav and Xu, Ming and Milford, Michael and Fischer, Tobias},
  booktitle=CVPR,
  pages={14141--14152},
  year={2021}
}

@article{DBLP:journals/ijcv/Lowe04,
  author    = {David G. Lowe},
  title     = {Distinctive Image Features from Scale-Invariant Keypoints},
  journal   = {Int. J. Comput. Vis.},
  volume    = {60},
  number    = {2},
  pages     = {91--110},
  year      = {2004}
}

@inproceedings{DBLP:conf/eccv/LiSH10,
  author    = {Yunpeng Li and
               Noah Snavely and
               Daniel P. Huttenlocher},
  title     = {Location Recognition Using Prioritized Feature Matching},
  booktitle = ECCV,
  pages     = {791--804},
  year      = {2010}
}

@misc{PoseLib,
  title = {{PoseLib - Minimal Solvers for Camera Pose Estimation}},
  author = {Viktor Larsson},
  URL = {https://github.com/vlarsson/PoseLib},
  year = {2020}
}

@inproceedings{DBLP:conf/cvpr/IrscharaZFB09,
  author    = {Arnold Irschara and
               Christopher Zach and
               Jan{-}Michael Frahm and
               Horst Bischof},
  title     = {From structure-from-motion point clouds to fast location recognition},
  booktitle = CVPR,
  pages     = {2599--2606},
  year      = {2009}
}

@article{peng2021megloc,
  title={Megloc: A robust and accurate visual localization pipeline},
  author={Peng, Shuxue and He, Zihang and Zhang, Haotian and Yan, Ran and Wang, Chuting and Zhu, Qingtian and Liu, Xiao},
  journal={arXiv preprint arXiv:2111.13063},
  year={2021}
}

@InProceedings{focustune,
    author    = {Nguyen, Son Tung and Fontan, Alejandro and Milford, Michael and Fischer, Tobias},
    title     = {FocusTune: Tuning Visual Localization Through Focus-Guided Sampling},
    booktitle = WACV,
    year      = {2024},
    pages     = {3606-3615}
}

@inproceedings{sattler2011fast,
  title={Fast image-based localization using direct 2d-to-3d matching},
  author={Sattler, Torsten and Leibe, Bastian and Kobbelt, Leif},
  booktitle=ICCV,
  pages={667--674},
  year={2011},
}

@InProceedings{Izquierdo_CVPR_2024_SALAD,
    author    = {Izquierdo, Sergio and Civera, Javier},
    title     = {Optimal Transport Aggregation for Visual Place Recognition},
    booktitle = CVPR,
    year      = {2024},
    pages = {17658--17668}
}

@inproceedings{DBLP:conf/cvpr/ShottonGZICF13,
  author    = {Jamie Shotton and
               Ben Glocker and
               Christopher Zach and
               Shahram Izadi and
               Antonio Criminisi and
               Andrew W. Fitzgibbon},
  title     = {Scene Coordinate Regression Forests for Camera Relocalization in {RGB-D}
               Images},
  booktitle = CVPR,
  pages     = {2930--2937},
  year      = {2013}
}

@inproceedings{DBLP:conf/cvpr/ValentinNSFIT15,
  author    = {Julien P. C. Valentin and
               Matthias Nie{\ss}ner and
               Jamie Shotton and
               Andrew W. Fitzgibbon and
               Shahram Izadi and
               Philip H. S. Torr},
  title     = {Exploiting uncertainty in regression forests for accurate camera relocalization},
  booktitle = CVPR,
  pages     = {4400--4408},
  year      = {2015}
}

@article{DBLP:journals/pami/BrachmannR22,
  author    = {Eric Brachmann and
               Carsten Rother},
  title     = {Visual Camera Re-Localization From {RGB} and {RGB-D} Images Using
               {DSAC}},
  journal   = PAMI,
  volume    = {44},
  number    = {9},
  pages     = {5847--5865},
  year      = {2022}
}

@inproceedings{hloc,
  author    = {Paul{-}Edouard Sarlin and
               Cesar Cadena and
               Roland Siegwart and
               Marcin Dymczyk},
  title     = {From Coarse to Fine: Robust Hierarchical Localization at Large Scale},
  booktitle = CVPR,
  pages     = {12716--12725},
  year      = {2019}
}

@inproceedings{active_search,
  author    = {Torsten Sattler and
               Bastian Leibe and
               Leif Kobbelt},
  title     = {Improving Image-Based Localization by Active Correspondence Search},
  booktitle = ECCV,
  pages     = {752--765},
  year      = {2012}
}

@inproceedings{huang2021vs,
  title={Vs-net: Voting with segmentation for visual localization},
  author={Huang, Zhaoyang and Zhou, Han and Li, Yijin and Yang, Bangbang and Xu, Yan and Zhou, Xiaowei and Bao, Hujun and Zhang, Guofeng and Li, Hongsheng},
  booktitle=CVPR,
  pages={6101--6111},
  year={2021}
}

@inproceedings{li2020hierarchical,
  title={Hierarchical scene coordinate classification and regression for visual localization},
  author={Li, Xiaotian and Wang, Shuzhe and Zhao, Yi and Verbeek, Jakob and Kannala, Juho},
  booktitle=CVPR,
  pages={11983--11992},
  year={2020}
}

@inproceedings{tang2023neumap,
  title={NeuMap: Neural Coordinate Mapping by Auto-Transdecoder for Camera Localization},
  author={Tang, Shitao and Tang, Sicong and Tagliasacchi, Andrea and Tan, Ping and Furukawa, Yasutaka},
  booktitle=CVPR,
  pages={929--939},
  year={2023}
}

@inproceedings{do2022learning,
  title={Learning to detect scene landmarks for camera localization},
  author={Do, Tien and Miksik, Ondrej and DeGol, Joseph and Park, Hyun Soo and Sinha, Sudipta N},
  booktitle=CVPR,
  pages={11132--11142},
  year={2022}
}

@inproceedings{sattler2018benchmarking,
  title={Benchmarking 6dof outdoor visual localization in changing conditions},
  author={Sattler, Torsten and Maddern, Will and Toft, Carl and Torii, Akihiko and Hammarstrand, Lars and Stenborg, Erik and Safari, Daniel and Okutomi, Masatoshi and Pollefeys, Marc and Sivic, Josef and others},
  booktitle=CVPR,
  pages={8601--8610},
  year={2018}
}

@inproceedings{brachmann2023accelerated,
  title={Accelerated Coordinate Encoding: Learning to Relocalize in Minutes using RGB and Poses},
  author={Brachmann, Eric and Cavallari, Tommaso and Prisacariu, Victor Adrian},
  booktitle=CVPR,
  pages={5044--5053},
  year={2023}
}

@inproceedings{brachmann2016uncertainty,
  title={Uncertainty-driven 6d pose estimation of objects and scenes from a single rgb image},
  author={Brachmann, Eric and Michel, Frank and Krull, Alexander and Yang, Michael Ying and Gumhold, Stefan and others},
  booktitle=CVPR,
  pages={3364--3372},
  year={2016}
}

@inproceedings{brachmann2018learning,
  title={Learning less is more-6d camera localization via 3d surface regression},
  author={Brachmann, Eric and Rother, Carsten},
  booktitle=CVPR,
  pages={4654--4662},
  year={2018}
}

@inproceedings{brachmann2019expert,
  title={Expert sample consensus applied to camera re-localization},
  author={Brachmann, Eric and Rother, Carsten},
  booktitle=ICCV,
  pages={7525--7534},
  year={2019}
}

@inproceedings{sarlin21pixloc,
  title={Back to the feature: Learning robust camera localization from pixels to pose},
  author={Sarlin, Paul-Edouard and Unagar, Ajaykumar and Larsson, Mans and Germain, Hugo and Toft, Carl and Larsson, Viktor and Pollefeys, Marc and Lepetit, Vincent and Hammarstrand, Lars and Kahl, Fredrik and others},
  booktitle=CVPR,
  pages={3247--3257},
  year={2021}
}

@inproceedings{rau2020predicting,
  title={Predicting visual overlap of images through interpretable non-metric box embeddings},
  author={Rau, Anita and Garcia-Hernando, Guillermo and Stoyanov, Danail and Brostow, Gabriel J and Turmukhambetov, Daniyar},
  booktitle=ECCV,
  pages={629--646},
  year={2020},
}

@inproceedings{wang2024dust3r,
  title={Dust3r: Geometric 3d vision made easy},
  author={Wang, Shuzhe and Leroy, Vincent and Cabon, Yohann and Chidlovskii, Boris and Revaud, Jerome},
  booktitle=CVPR,
  pages={20697--20709},
  year={2024}
}

@inproceedings{wang2025vggt,
  title={Vggt: Visual geometry grounded transformer},
  author={Wang, Jianyuan and Chen, Minghao and Karaev, Nikita and Vedaldi, Andrea and Rupprecht, Christian and Novotny, David},
  booktitle=CVPR,
  pages={5294--5306},
  year={2025}
}

@inproceedings{reloc3r,
  title={Reloc3r: Large-scale training of relative camera pose regression for generalizable, fast, and accurate visual localization},
  author={Dong, Siyan and Wang, Shuzhe and Liu, Shaohui and Cai, Lulu and Fan, Qingnan and Kannala, Juho and Yang, Yanchao},
  booktitle=CVPR,
  pages={16739--16752},
  year={2025}
}

@inproceedings{deviloc,
  title={Learning to Produce Semi-dense Correspondences for Visual Localization},
  author={Giang, Khang Truong and Song, Soohwan and Jo, Sungho},
  booktitle=CVPR,
  year={2024}
}

@inproceedings{weinzaepfel2023croco,
  title={Croco v2: Improved cross-view completion pre-training for stereo matching and optical flow},
  author={Weinzaepfel, Philippe and Lucas, Thomas and Leroy, Vincent and Cabon, Yohann and Arora, Vaibhav and Br{\'e}gier, Romain and Csurka, Gabriela and Antsfeld, Leonid and Chidlovskii, Boris and Revaud, J{\'e}r{\^o}me},
  booktitle=ICCV,
  pages={17969--17980},
  year={2023}
}

@inproceedings{mast3r,
  title={Grounding image matching in 3d with mast3r},
  author={Leroy, Vincent and Cabon, Yohann and Revaud, J{\'e}r{\^o}me},
  booktitle=ECCV,
  pages={71--91},
  year={2024},
}

@inproceedings{yang2022scenesqueezer,
  title={Scenesqueezer: Learning to compress scene for camera relocalization},
  author={Yang, Luwei and Shrestha, Rakesh and Li, Wenbo and Liu, Shuaicheng and Zhang, Guofeng and Cui, Zhaopeng and Tan, Ping},
  booktitle=CVPR,
  pages={8259--8268},
  year={2022}
}

@inproceedings{cheng2019cascaded,
  title={Cascaded parallel filtering for memory-efficient image-based localization},
  author={Cheng, Wentao and Lin, Weisi and Chen, Kan and Zhang, Xinfeng},
  booktitle=ICCV,
  pages={1032--1041},
  year={2019}
}

@inproceedings{dusmanu2019d2,
  title={D2-net: A trainable cnn for joint description and detection of local features},
  author={Dusmanu, Mihai and Rocco, Ignacio and Pajdla, Tomas and Pollefeys, Marc and Sivic, Josef and Torii, Akihiko and Sattler, Torsten},
  booktitle=CVPR,
  pages={8092--8101},
  year={2019}
}

@article{revaud2019r2d2,
  title={R2d2: Reliable and repeatable detector and descriptor},
  author={Revaud, Jerome and De Souza, Cesar and Humenberger, Martin and Weinzaepfel, Philippe},
  journal=NIPS,
  volume={32},
  year={2019}
}

@inproceedings{rscore,
  title={{R-SCoRe}: Revisiting Scene Coordinate Regression for Robust Large-Scale Visual Localization},
  author={Jiang, Xudong and Wang, Fangjinhua and Galliani, Silvano and Vogel, Christoph and Pollefeys, Marc},
	booktitle = CVPR,
  year={2025},
pages = {11536--11546}
}
}
\end{document}